\documentclass[letterpaper, 10 pt, conference]{ieeeconf}

\IEEEoverridecommandlockouts                              % This command is only needed if 
\usepackage{graphics}
\usepackage{graphicx}
\usepackage{amssymb}
\usepackage{amsmath}
\usepackage{epsfig}
\usepackage{cite}
\usepackage{algorithm}
\usepackage{algpseudocode}
\usepackage{comment}

\title{\LARGE \bf
Staged Contact Optimization: Combining Contact-Implicit and Multi-Phase Hybrid Trajectory Optimization}
        
\author{Michael~R.~Turski,
        Joseph~Norby,
        and~Aaron~M.~Johnson%
\thanks{* This work was supported by the National Science Foundation
    under grant \#DGE-1745016 as well as the U.S. Army Research
    Office under grant \#W911NF-19-1-0080. The views and conclusions contained
    in this document are those of the authors and should not be interpreted as
    representing the official policies, either expressed or implied, of the National Science Foundation,
    the Army Research Office, or the U.S. Government.
    The U.S. Government is authorized to reproduce and distribute reprints
    for Government purposes notwithstanding any copyright notation herein.}%
\thanks{Department of Mechanical Engineering, Carnegie Mellon University, Pittsburgh, PA 15213, USA, amj1@andrew.cmu.edu.}}%

\begin{document}
\bstctlcite{IEEEexample:BSTcontrol}

\maketitle
\thispagestyle{empty}
\pagestyle{empty}

%%%%%%%%%%%%%%%%%%%%%%%%%%%%%%%%%%%%%%%%%%%%%%%%%%%%%%%%%%%%%%%%%%%%%%
\begin{abstract}
Trajectory optimization problems for legged robots are commonly formulated with fixed contact schedules. These multi-phase Hybrid Trajectory Optimization (HTO) methods result in locally optimal trajectories, but the result depends heavily upon the predefined contact mode sequence. Contact-Implicit Optimization (CIO) offers a potential solution to this issue by allowing the contact mode to be determined throughout the trajectory by the optimization solver. However, CIO suffers from long solve times and convergence issues. This work combines the benefits of these two methods into one algorithm: Staged Contact Optimization (SCO). SCO tightens constraints on contact in stages, eventually fixing them to allow robust and fast convergence to a feasible solution. Results on a planar biped and spatial quadruped demonstrate speed and optimality improvements over CIO and HTO. These properties make SCO well suited for offline trajectory generation or as an effective tool for exploring the dynamic capabilities of a robot.
\end{abstract}

%\begin{keywords} %May have to switch to IEEEkeywords for ieeetrans
%Multi-Contact Whole-Body Motion Planning and Control, Legged Robots, Optimization and Optimal Control
%\end{keywords}

%%%%%%%%%%%%%%%%%%%%%%%%%%%%%%%%%%%%%%%%%%%%%%%%%%%%%%%%%%%%%%%%%%%%%%
\section{Introduction}
Trajectory optimization is a common method of determining optimal behaviors for legged robot systems, solving for body and joint state trajectories with feed-forward motor torques, often written as a nonlinear programming problem (NLP)~\cite{kelly2017introduction}. Standard trajectory optimization problems are set up for continuous systems. For legged robots, the problem becomes much more complex due to their inherently nonsmooth impact dynamics, typically modeled as a hybrid dynamical system (i.e.\ a system with both continuous and discrete states)~\cite{hybridsystems,johnson2016hybrid}.
In order to locomote, legged robots impact and push off of the ground changing their discrete contact mode. This changes their dynamics and creates discontinuities in velocity that cannot be directly handled in a standard trajectory optimization problem. 

\begin{figure}[t]
\begin{center}
\includegraphics[width=0.9\columnwidth]{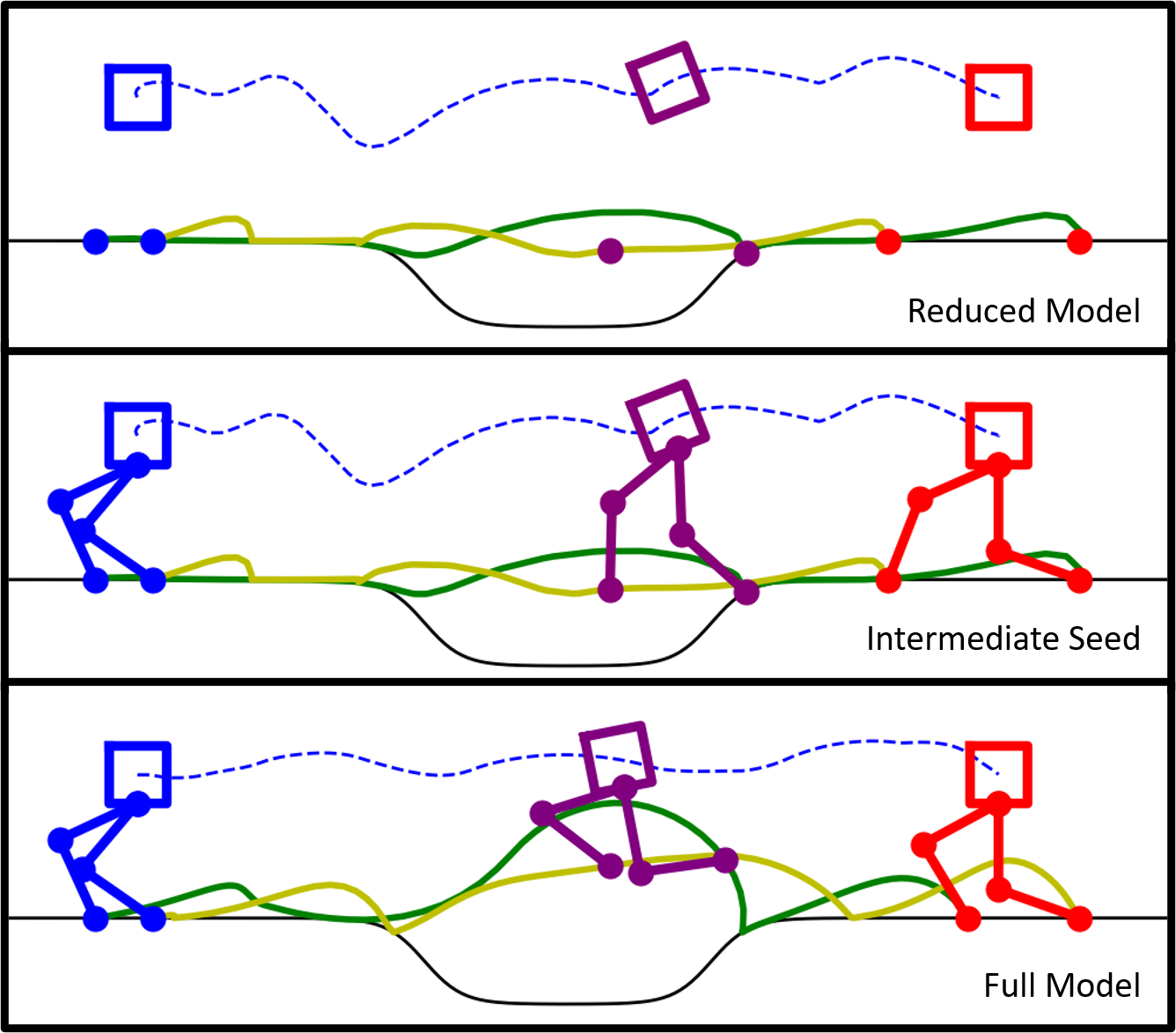}
\end{center}
\vspace{-1em}
\caption{Staged Contact Optimization (SCO) approach applied to a planar biped robot crossing a gap in the floor. Top: A reduced order model is optimized with CIO to determine the optimal contact sequence. Middle: The full robot model is used to fill in lost information. Bottom: The full robot model is optimized with HTO to determine a refined behavior.}
\label{Abstract_Picture} 
\end{figure}

Trajectory optimization for hybrid systems is commonly formulated with fixed mode schedules. This method, commonly known as multi-phase or Hybrid Trajectory Optimization (HTO)~\cite{kelly2017introduction,von1999user}, links different continuous phases together with discrete reset events in between~\cite{schultz2009modeling,posa2016optimization}. This results in a locally optimal trajectory with good convergence by avoiding the combinatorial nature of hybrid events \cite{buss2000towards}. However, the result relies heavily upon the predefined contact mode sequence, which can limit the performance and capability of the robot. For example, it may be beneficial to switch from a walk to a jump to cross a large gap (Fig.~\ref{Abstract_Picture}), rather than using a running gait for the entire behavior. Problem complexity, caused either by the robot model's complexity or the difficulty of the desired behavior, causes the ideal contact mode sequence to be unknown a priori in many cases.

Contact-Implicit Trajectory Optimization (CIO)
%, also known as contact-invariant optimization or optimization through contact, 
offers a solution to this issue by allowing the contact mode to be determined throughout the trajectory~\cite{posa2014direct,mordatch2012discovery}. This allows the optimization to make and break contact as needed, but introduces numerical complexity which can lead to long computation times and poor convergence.

\begin{figure*}[t]
\begin{center}
\includegraphics[width=.9\textwidth]{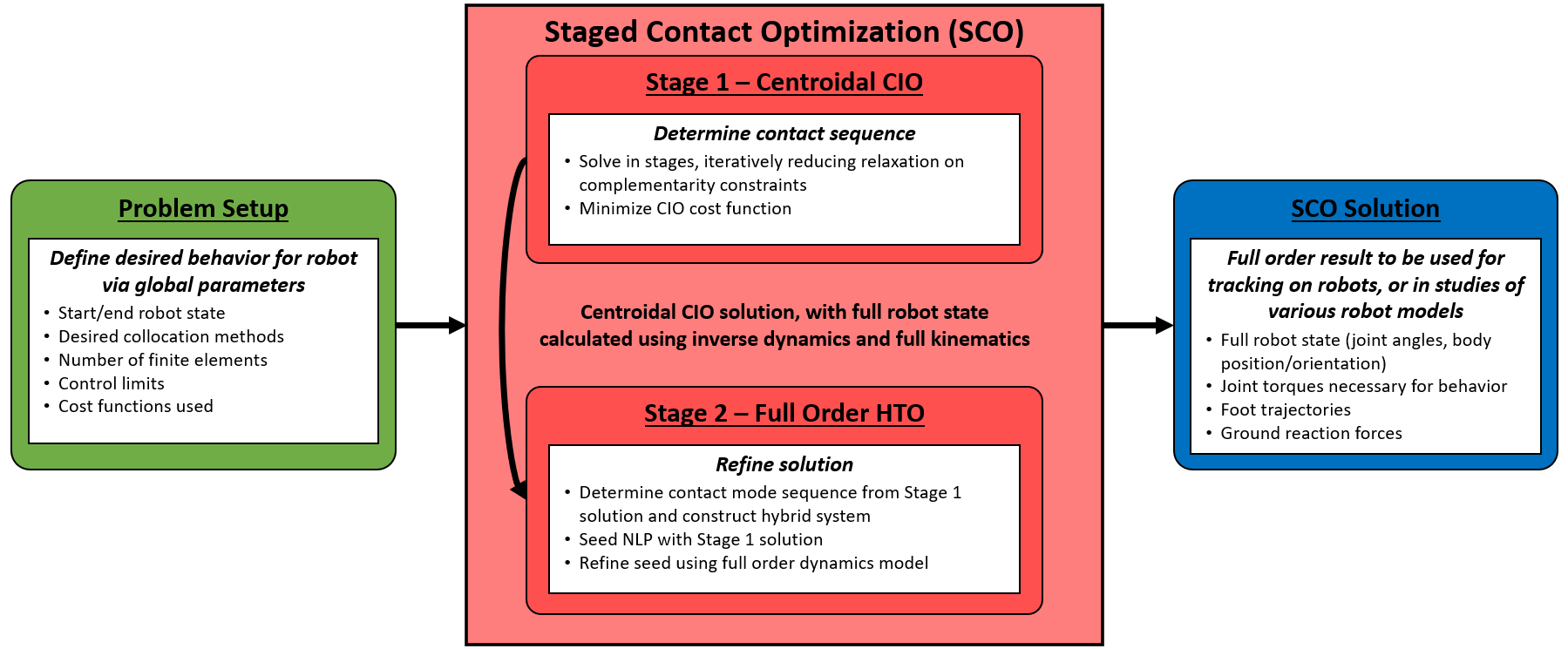}
\end{center}
\vspace{-1em}
\caption{Overview of SCO algorithm. CIO is used on a centroidal dynamics model to determine the contact mode sequence for the desired task, as well as a good initial guess for the body and feet trajectories. The full-order robot model is used to determine the missing information lost by using the reduced order model, the contact mode is fixed, and HTO is used to refine the behavior.}
\label{SCO_FlowChart} 
\end{figure*}

This work combines CIO and HTO into one algorithm -- Staged Contact-Implicit and Hybrid Trajectory Optimization, simplified as Staged Contact Optimization (SCO) -- which takes advantage of the benefits of each. SCO solves motion planning problems by leveraging a centroidal dynamics model with CIO and iteratively tightening the complementarity constraints before fixing the contact sequence and deploying HTO to resolve the full-order cost function and constraints, Fig.~\ref{SCO_FlowChart}. This staged approach harnesses CIO's ability to reason about contact modes while avoiding its poor convergence, then takes on HTO's speed and convergence properties for efficient full-order refinement. SCO is able to create dynamic behaviors for a wide range of complex situations and can find behaviors that are more optimal than HTO with fixed gait cycles in a similar amount of time. 

The paper is organized as follows: Section \ref{sec:related} gives an overview of some related work. Section \ref{sec:method} describes the algorithm, including subsections describing the centroidal CIO and HTO portions of the algorithm. Section \ref{sec:performance_comparison} presents experimental studies, comparing SCO against full-order CIO, hierarchical CIO (HCIO), and standard HTO on planar 5-link biped and spatial 18 DOF quadruped models. Finally, Section \ref{sec:conclusion} concludes with a discussion and potential future work.

\subsection{Related Works}
\label{sec:related}

One common approach to optimizing through contact is to simplify the underlying system or constraints while still optimizing the contact sequence and trajectory simultaneously. Some works use reduced-order models such as centroidal dynamics with heuristics to promote feasibility on the full system~\cite{dai2014whole,winkler2018gait,ponton2021efficient}. These have demonstrated improved convergence, but the approximations necessarily result in either overly conservative or optimistic behaviors. Other works have retained the full-order dynamics of the system but relaxed the contact constraints, by employing smooth contacts \cite{tassa2012synthesis,mordatch2012discovery}, spring-damper models \cite{neunert2018whole}, or by moving the contact constraint violation to the cost function \cite{neunert2016efficient,todorov2011convex,manchester2019contact}. These avoid discontinuous hard contact at the cost of allowing small forces away from contact surfaces or requiring very small timesteps. Some recent works have handled both full-order dynamics and hard contact by linearizing the dynamics about a trajectory, but as a result are more suitable for trajectory tracking than planning \cite{kong2021ilqr,le2021linear}.

Other methods focus on improving the algorithm which solves the NLP. One example is to employ Mixed Integer Programming (MIP), which encodes the contact mode as an integer optimization variable, providing contact information at the expense of combinatorial complexity \cite{ponton2021efficient,aceituno2017simultaneous,kuindersma2016optimization}. A different approach is to increase the accuracy of the solution, which allows for a coarser discretization of the trajectory. This has been achieved through variational integrators \cite{manchester2019contact} and higher-order collocation methods \cite{patel2019contact,chao2017step}, although they still remain quite slow and can struggle with convergence.

Some researchers have blended these approaches by combining solutions of relaxed and exact problems. One example is to first fix the complementarity constraint relaxation to a large value and then iteratively reduce it~\cite{hoheisel2013theoretical,posa2014direct}. Another approach first solves the problem with first-order collocation and a large relaxation and then uses this solution to warm start a problem with third-order collocation and a small relaxation \cite{patel2019contact}. 
Others use centroidal dynamics to inform a full-order dynamics model solve with CIO \cite{mastalli2016hierarchical}.
These methods ensure that the final result will meet the physical constraints of the problem and result in overall decreases in solve times, but still struggle with convergence due to the full-order complexity of the dynamics in combination with the complementarity constraints. More recently, MIP was applied to an iterative convex approximation of centroidal dynamics and full kinematics, achieving impressive convergence rates~\cite{ponton2021efficient}. Similar to SCO, this method decomposes the problem into two sub-problems, contact surface selection and motion planning. However, SCO solves these two problems in its first CIO step, and then extends to solve the full-order problem.

%%%%%%%%%%%%%%%%%%%%%%%%%%%%%%%%%%%%%%%%%%%%%%%%%%%%%%%%%%%%%%%%%%%%%%
\section{Method} \label{sec:method}

This paper presents Staged Contact Optimization (SCO), Fig.~\ref{SCO_FlowChart}.
The algorithm has two main goals: 1) allow the contact sequence to vary, so that SCO can find results without a pre-defined contact sequence. 2) Provide a full-order optimization result for a given system and behavior. To meet these requirements, CIO is used in the first stage of SCO to address the first goal (Sec.~\ref{sec:centroidalCIO}), while HTO is used in the final stage to address the second goal (Sec.~\ref{sec:HTO}). 

In SCO, dynamics and control inputs are transcribed into an NLP which is solved using standard solvers. The optimization is formulated using direct methods, specifically direct collocation \cite{kelly2017introduction}. Both first-order backward Euler collocation and third-order Radau collocation \cite{patel2019contact,beigler2010nonlinearprogramming} are used to integrate the robot states and the cost functions.

The first-order method applied to a generic variable $z$ is:
\begin{equation} \label{eq:euler}
    z_{n} = z_{n-1} + h\dot{z}_{n}
\end{equation}
where $n$ is the index of the finite elements and $h$ is the time step duration. The third-order Radau collocation method is:
\begin{align} %\label{eq:radau}
    z_{n,p} = z_{n-1,2} + \eta_{n,p}, \quad
    \eta_{n,p} &= h\sum_{i=0}^2 \Gamma[p,i]\dot{z}_{n,i},  \quad % &\forall p \in \{0,1,2\}  \\
    %, \quad &\forall p \in \{0,1,2\}  %\\
%    \Gamma &= \begin{bmatrix}
%    \frac{440-35\sqrt{6}}{1800} & \frac{296-169\sqrt{6}}{1800} & \frac{-2+3\sqrt{6}}{225}\\
%    \frac{296+169\sqrt{6}}{1800} & \frac{440+35\sqrt{6}}{1800} & \frac{-2-3\sqrt{6}}{225}\\
%    \frac{16-\sqrt{6}}{36} & \frac{16+\sqrt{6}}{36} & \frac{1}{9}\\
%    \end{bmatrix} 
\label{eq:radau_end}
    \end{align}
where $\eta$ is a slack variable, $p$ is the collocation point, and $\Gamma$ are constants defined for third-order Radau collocation, \cite{beigler2010nonlinearprogramming}. Both first and third-order are used in SCO and the usage of each are discussed in the definition of the stages of SCO.

%%%%%%%%%%%%%%%%%%%%%%%%%%%%%%%%%%%%%%%%%%%%%%%%%%%%%%%%%%%%%%%%%%%%%%
\subsection{Centroidal CIO Stage}
\label{sec:centroidalCIO}
%%%%%%%%%%%%%%%%%%%%%%%%%%%%%%%%%%%%%%%%%%%%%%%%%%%%%%%%%%%%%%%%%%%%%%
\subsubsection{Dynamics}

For the first portion of SCO, CIO is used on a centroidal dynamics model of the robot:
\begin{align} \label{eq:cio_dynamics}
    m \ddot{\mathbf{r}} &= \sum_{i} \mathbf{\lambda}_i + m\mathbf{g}, \qquad
    \mathbf{I} \ddot{\mathbf{\phi}} = \sum_{i} (\mathbf{a}_i - \mathbf{r}) \times \mathbf{\lambda}_i
\end{align}
where $m$ is the mass of the robot, $\mathbf{I} \in \mathbb{R}^3$ is the inertia matrix, $\mathbf{r} \in \mathbb{R}^3$ is the position of the robot's center of mass (COM), $\dot{\phi} \in \mathbb{R}^3$ is the angular velocity vector, $\mathbf{\lambda}_i \in \mathbb{R}^3$ is the ground reaction force/impulse vector at the $i^{th}$ contact point located at $\mathbf{a}_i \in \mathbb{R}^3$, and $\mathbf{g} \in \mathbb{R}^3$ is the gravity vector.

Despite not including the joint information within the dynamics, full-order kinematics are used to constrain joint angles variables, $\theta$, (subject to limits $\theta_l \leq \theta \leq \theta_u$) to the COM position, $\mathbf{r}$, and orientation, $\varphi$, as well as the contact point locations through the forward kinematics, $\mathbf{g}_i$, for the $i^{th}$ foot, 
\begin{align}
    \mathbf{a}_i = \mathbf{g}_i(\mathbf{q}), \quad \mathbf{q} = \begin{bmatrix}\theta\\ \mathbf{r}\\ \varphi \end{bmatrix}, \quad \theta_l \leq \theta \leq \theta_u
    \label{eq:fk_a}
\end{align}
Collocation constraints are applied to the contact point positions and velocities to enforce that the contact point trajectories are smooth.

%%%%%%%%%%%%%%%%%%%%%%%%%%%%%%%%%%%%%%%%%%%%%%%%%%%%%%%%%%%%%%%%%%%%%%
\subsubsection{Complementarity and Friction}

Complementarity constraints are added to enforce the non-penetration of contacts:
\begin{equation} \label{eq:non_penetration}
    \lambda_{N,i} h_i(\mathbf{a}_i) = 0, \quad \lambda_{N,i} \geq 0, \quad h_i(\mathbf{a}_i) \geq 0 
\end{equation}
where $\lambda_{N,i}$ is the normal force and $h_i$ is the distance between the contact point and the terrain. Additional constraints can be added to enforce the friction cone, as in  \cite{posa2014direct}. A standard implementation of CIO allows for static and sliding friction, however, this work only considers static friction as allowing for sliding friction can cause difficulties implementing the solutions in simulations and in real-world applications. The static friction cone is enforced by:
\begin{align} \label{eq:friction_comp}
    &\lambda_{T_j,i} \gamma_i(\dot{\mathbf{a}}_i) = 0, \quad \mu \lambda_{N,i} - \sum_j\lambda_{T_j,i} \geq 0, \quad \lambda_{T_j,i} \geq 0
\end{align}
where $\gamma_i$ is the relative tangential velocity and $\lambda_{T_j,i}$ are the tangential force components.

%%%%%%%%%%%%%%%%%%%%%%%%%%%%%%%%%%%%%%%%%%%%%%%%%%%%%%%%%%%%%%%%%%%%%%
\subsubsection{Third-Order Slack Variables for Complementarity Constraints}

For first-order collocation based approaches, the complementarity constraints in \eqref{eq:non_penetration} and \eqref{eq:friction_comp} can be implemented directly as written for each time-step. However, for higher-order collocation, these constraints must be considered across the collocation points within a finite element. This is done by enforcing that the contact mode must not change within the finite element. Following the process in \cite{patel2019contact}, the following slack variables are added to the NLP:
\begin{align} \label{eq:cio_slacks}
    \alpha^{\prime}_i = \sum_{p=0}^2 \alpha^{\prime}_{i,p}, \quad
    \beta^{\prime}_i = \sum_{p=0}^2 \beta^{\prime}_{i,p}, \quad
    \alpha^{\prime}_i \beta^{\prime}_i = 0
\end{align}
where $\alpha^{\prime}_{i,p}$ and $\beta^{\prime}_{i,p}$ are set equal to their respective variables at the $p^{th}$ collocation point per $i^{th}$ finite element in equations \eqref{eq:non_penetration} and \eqref{eq:friction_comp}.
The complementarity constraints can be simplified using a relaxation \cite{hoheisel2013theoretical} as follows:
\begin{equation} \label{eq:cio_relaxed}
    \alpha^{\prime}_i \beta^{\prime}_i \leq \epsilon
\end{equation}

%%%%%%%%%%%%%%%%%%%%%%%%%%%%%%%%%%%%%%%%%%%%%%%%%%%%%%%%%%%%%%%%%%%%%%
\subsubsection{Problem Setup}

As with any trajectory optimization problem setup, initial and final robot states are applied based on the desired behavior and a cost function is selected to shape the solution. Commonly, the cost is a function of the energy input into the system, which requires access to the joint torques. Using centroidal CIO creates a unique difficulty in this regard, as the torques are no longer decision variables. Therefore, the joint torques, $\mathbf{\tau}$, must be approximated using the Jacobian of the contact point kinematics, $\mathbf{g}$, \eqref{eq:fk_a},
\begin{align} \label{eq:contact_jacobian}
    \mathbf{\tau} \approx \mathbf{A}^T \mathbf{\lambda},
    \quad
    \mathbf{A} = \frac{\partial \mathbf{g(\mathbf{q})}}{\partial \mathbf{q}}
\end{align}
This approximation is added as a constraint within the NLP so that the joint torques can be included in the cost function and control limits can be placed on them if desired. 

%%%%%%%%%%%%%%%%%%%%%%%%%%%%%%%%%%%%%%%%%%%%%%%%%%%%%%%%%%%%%%%%%%%%%%
%\subsubsection{Cost Function}
Often, simply minimizing the integral of the torque input squared will result in quality behaviors.
However, in order to find a result with this centroidal model that approaches a solution found with full-order dynamics, additional terms are used to discourage unwanted movements in the trajectory:
% \begin{equation} \label{eq:cio_cost}
%     c(\mathbf{z}) = \int_0^{t_f} (\mathbf{\tau}^T\mathbf{\tau}\, +\, \ddot{\mathbf{a}}^T\ddot{\mathbf{a}}\, + \,\ddot{\mathbf{q}}^T \ddot{\mathbf{q}}\, +\, (\mathbf{q} - \mathbf{q_{0}})^T(\mathbf{q} - \mathbf{q_{0}})) \text{dt}
% \end{equation}
\begin{equation} \label{eq:cio_cost}
    c(\mathbf{z}) = \int_0^{t_f} (\mathbf{\tau}^T\mathbf{\tau}\, +\, \ddot{\mathbf{a}}^T\ddot{\mathbf{a}}\, + \,\ddot{\mathbf{q}}^T \ddot{\mathbf{q}}\, +\, \Delta\mathbf{q_{0}}^T\Delta\mathbf{q_{0}}) \text{dt}
\end{equation}
where $\Delta\mathbf{q_{0}}$ is the deviation from the nominal state of the robot and $\mathbf{z}$ contains all decision variables. Weighting this cost function may be beneficial in some cases.  This cost function, based on \cite{dai2014whole}, penalizes the estimated joint torques to minimize energy usage, foot and joint accelerations to ensure smooth swing trajectories, and deviations from the nominal robot posture to avoid unwanted joint configurations.

%%%%%%%%%%%%%%%%%%%%%%%%%%%%%%%%%%%%%%%%%%%%%%%%%%%%%%%%%%%%%%%%%%%%%%
\subsubsection{Solving Centroidal CIO}

The optimization problem for the Centroidal CIO stage is thus:
\begingroup
\allowdisplaybreaks
\begin{align*} \label{problem:centroidal_cio} 
\textbf{NLP1: Centroidal}&\textbf{ CIO} \\
    \textbf{Minimize:} &
    \quad  \text{Centroidal cost function \eqref{eq:cio_cost} }\\
    \textbf{With respect to:} \\
    r_{n,p}, \dot{r}_{n,p}, \ddot{r}_{n,p} \quad & \text{COM position, velocity, acceleration} \\
    \varphi_{n,p}, \dot{\phi}_{n,p}, \ddot{\phi}_{n,p} \quad & \text{COM orientation, velocity, acceleration} \\
    a_{n,p}, \dot{a}_{n,p}, \ddot{a}_{n,p} \quad & \text{Foot positions, velocities, accelerations} \\
    \theta_n, \dot{\theta}_n, \tau_{n} \quad & \text{Joint angles, velocities, and torques} \\
    \lambda_{n,p} \quad & \text{Ground reaction forces/impulses} \\
    \eta_{n,p} \quad & \text{Collocation slack variables} \\
    \alpha_{n,p}, \beta_{n,p}, \alpha^{'}_{n}, \beta^{'}_{n} \quad & \text{Complementarity slack variables} \\
    \textbf{Subject to:} \\
    \eqref{eq:euler}\text{ or }\eqref{eq:radau_end} \quad & \text{Collocation} \\
    \eqref{eq:cio_dynamics} \quad & \text{Centroidal dynamics} \\
    \eqref{eq:fk_a} \quad & \text{Forward kinematics}\\
    \eqref{eq:non_penetration}-\eqref{eq:cio_relaxed} \quad & \text{Contact \& complementarity constraints} \\
    \eqref{eq:contact_jacobian} \quad & \text{Torque Approximation constraint} 
\end{align*}
\endgroup
Where the decision variables are indexed by finite element, $n$, and collocation point, $p$.
Also included are problem specific constraints, such as start and end conditions, time constraints, control limits, and path constraints. 

In order to quickly and consistently solve the centroidal CIO NLP, an iterative solving approach is used. Algorithm~\ref{algo:centroidal_cio_flat} shows the details of this approach, depending on whether the final result should use first-order or third-order collocation.

\begin{algorithm}[t] 
\caption{SCO Stage 1 - Centroidal CIO}
\begin{algorithmic}

    \State Define centroidal CIO variables and constraints

    \State Solve NLP1 w/ loose $\epsilon$, coarse FE, $1^{st}$ order
    \State Solve NLP1 w/ loose $\epsilon$, fine FE, $1^{st}$ order
    
    \If{Use Terrain Map = True}
        \State Activate terrain map constraints
        \State Solve NLP1 w/ loose $\epsilon$, fine FE, $1^{st}$ order
    \EndIf
    
    \If{Desired Collocation = $1^{st}$}
    
        \State Solve NLP1 w/ moderate $\epsilon$, fine FE, $1^{st}$ order
        \State Solve NLP1 w/ tight $\epsilon$, fine FE, $1^{st}$ order
        
    \ElsIf{Desired Collocation = $3^{rd}$}
    
        \State Solve NLP1 w/ loose $\epsilon$, fine FE, $3^{rd}$ order
        \State Solve NLP1 w/ moderate $\epsilon$, fine FE, $3^{rd}$ order
        \State Solve NLP1 w/ tight $\epsilon$, fine FE, $3^{rd}$ order
        
    \EndIf
    
\end{algorithmic}\label{algo:centroidal_cio_flat}
\end{algorithm}

Here we utilize several strategies from previous works, including mesh refinement and iteratively decreasing the complementarity relaxation. When iterating through these steps, all variables in the NLP are initially set to the result of the previous solution. In the case of an increase in finite elements (FE), mesh refinement is used to interpolate the results in order to best seed the next solve. When transitioning from first-order to third-order collocation, the collocation points are initialized by forward solving the collocation problem.

The decision between using first-order and third-order collocation in the centroidal CIO stage is made based on problem specifics and desired results. In our experience, first-order is preferable if faster solve times are desired. Using third-order in this step often results in smoother results for the centroidal CIO step, but takes more time to solve.

In the case of varying height terrain, a slightly different solving strategy is used to have better results. The first two calls in Algorithm~\ref{algo:centroidal_cio_flat} are made assuming flat terrain. The second call is then repeated adding in the height map, and all remaining calls use the actual height map.

%%%%%%%%%%%%%%%%%%%%%%%%%%%%%%%%%%%%%%%%%%%%%%%%%%%%%%%%%%%%%%%%%%%%%%
\subsection{Full-Order HTO Stage}
\label{sec:HTO}

%%%%%%%%%%%%%%%%%%%%%%%%%%%%%%%%%%%%%%%%%%%%%%%%%%%%%%%%%%%%%%%%%%%%%%
\subsubsection{Dynamics}

For the second stage of SCO, the centroidal CIO result is refined using a full-order HTO. This requires a different setup from centroidal CIO. The full-order dynamics of the robot are (e.g.\ following \cite{murray2017mathematical,paper:johnson_selfmanip_2013}):
\begin{equation} \label{eq:full_dynamics}
    \mathbf{M(\mathbf{q})}\ddot{\mathbf{q}} + \mathbf{C(\mathbf{q},\dot{\mathbf{q}})}\dot{\mathbf{q}} + \mathbf{N(\mathbf{q})} + \mathbf{A(\mathbf{q})}^T \mathbf{\lambda} = \Upsilon(\tau) 
\end{equation}
where $\mathbf{M}$ is the mass matrix, $\mathbf{C}$ is the Coriolis matrix, $\mathbf{N}$ is the gravitational forces, and $\Upsilon$ is the input vector. 

%%%%%%%%%%%%%%%%%%%%%%%%%%%%%%%%%%%%%%%%%%%%%%%%%%%%%%%%%%%%%%%%%%%%%%
\subsubsection{Hybrid System}

The dynamics of the system, in particular $\mathbf{A}$, depend on the contact mode at a given time. In regular HTO, this contact sequence is defined by the user, however in SCO the contact sequence is extracted from the CIO result by looking at the contact forces over time. Any contact with a non-zero force in a given finite element is considered active. This is done automatically by the algorithm, allowing SCO to seamlessly transition from solving a CIO problem to a HTO problem.

Each contact mode phase is allocated 25 finite elements (a number proportional to the phase duration in the CIO solution would also be reasonable). We found that 25 FE has given good convergence properties and high resolution for phases up to around 3 seconds each. The time of each phase, $t_s$, is added as a decision variable and allowed to vary.

From the contact sequence, constraints are applied to enforce the contact mode on the corresponding finite elements. If the $i^{th}$ contact point is active, then a zero-velocity constraint is applied across the entire phase:
\begin{equation} \label{eq:hybrid_zero_vel}
    \mathbf{A}_i\dot{\mathbf{q}} = 0
\end{equation}
If the $i^{th}$ contact point is inactive, then a zero-contact force is applied as well as a non-penetration constraint:
\begin{equation} \label{eq:hybrid_zero_lam}
    \mathbf{\lambda}_i = 0, \qquad
    h_i(\mathbf{a}_i) \geq 0
\end{equation}
where $h_i$ is the distance to the terrain. Finally, if the $i^{th}$ contact point is active in phase $s$, the following constraint is applied to the last collocation point of the $(s-1)^{th}$ phase:
\begin{equation} \label{eq:hybrid_landing}
     h_{i}(\mathbf{a}_i)  = 0
\end{equation}

To link the continuous phases of each contact mode in the sequence, reset maps (impact or liftoff conditions) are applied between the last FE of the current phase and the first FE of the next phase:
\begin{align} \label{eq:hybrid_reset}
    \mathbf{q}^+ &= \mathbf{q}^-, \qquad
    \mathbf{M}(\dot{\mathbf{q}}^+-\dot{\mathbf{q}}^-) = \mathbf{A}^T \hat{\mathbf{\rho}}_s 
\end{align} 
where $\mathbf{q}^+$ and $\mathbf{q}^-$ are the state before and after the event and $\hat{\mathbf{\rho}}_s$ is the impulse from the event. These combined with \eqref{eq:hybrid_zero_vel} define the plastic impact law. 

%%%%%%%%%%%%%%%%%%%%%%%%%%%%%%%%%%%%%%%%%%%%%%%%%%%%%%%%%%%%%%%%%%%%%%
\subsubsection{Friction Cone}

Finally, the following constraints are applied to enforce the static friction cone, for both the contact forces and impulses:
\begin{align} \label{eq:hybrid_friction}
    \mu \lambda_{N,i} - \sum_j\lambda_{T_j,i} \geq 0 &\quad \lambda_{N,i} \geq 0&\quad \lambda_{T_j,i} \geq 0\\
    \mu \hat{\mathbf{\rho}}_{N,i} - \sum_j\hat{\mathbf{\rho}}_{T_j,i} \geq 0 &\quad \hat{\mathbf{\rho}}_{N,i} \geq 0  &\quad \hat{\mathbf{\rho}}_{T_j,i} \geq 0 \label{eq:hybrid_friction_end}
\end{align}
where $\hat{\mathbf{\rho}}_{N,i}$ is the normal impulse component and $\hat{\mathbf{\rho}}_{T_j,i}$ are the tangential impulse components at each contact point.

%%%%%%%%%%%%%%%%%%%%%%%%%%%%%%%%%%%%%%%%%%%%%%%%%%%%%%%%%%%%%%%%%%%%%%
\subsubsection{Cost Function}

To finish setting up the HTO NLP, a cost function must be selected. In the examples in this paper, a torque squared cost is used:
\begin{equation} \label{eq:hybrid_cost}
    c(\mathbf{z}) = \int_0^{t_f} \mathbf{\tau}^T\mathbf{\tau} \text{dt}
\end{equation}
By using full-order dynamics, the cost function has direct access to exact joint torques, avoiding need for additional terms in \eqref{eq:cio_cost}. More complexity can be added to the cost, such as minimizing the total mechanical work or the total energy input, however, these problems can be harder to solve.

%%%%%%%%%%%%%%%%%%%%%%%%%%%%%%%%%%%%%%%%%%%%%%%%%%%%%%%%%%%%%%%%%%%%%%
\subsubsection{Solving Full-Order HTO}

The optimization problem for the full order HTO stage is thus:
\begingroup
\allowdisplaybreaks
\begin{align*}
\textbf{NLP2: Full-order}&\textbf{ HTO} \\
    \textbf{Minimize:} &
    \quad  \text{Full-order cost function \eqref{eq:hybrid_cost} }\\
    \textbf{With respect to:} \\
    t_{s} \quad & \text{contact mode time duration} \\
    q_{s,n,p}, \dot{q}_{s,n,p}, \ddot{q}_{s,n,p}  \quad & \text{Robot state, velocity, acceleration} \\
    a_{s,n,p}, \dot{a}_{s,n,p} \quad & \text{Foot position, velocity} \\
    \tau_{s,n} \quad & \text{Joint torques} \\
    \lambda_{s,n,p},\hat{\rho}_{s} \quad & \text{Ground reaction forces and impulses} \\
    \eta_{s,n,p} \quad & \text{Collocation slack variables} \\
    \textbf{Subject to:} \\
    \eqref{eq:euler}\text{ or }\eqref{eq:radau_end} \quad & \text{Collocation} \\
    \eqref{eq:fk_a} \quad & \text{Forward kinematics}\\
    \eqref{eq:full_dynamics} \quad & \text{Full order dynamics} \\
    \eqref{eq:hybrid_zero_vel}-\eqref{eq:hybrid_reset} \quad & \text{Hybrid system constraints} \\
    \eqref{eq:hybrid_friction}-\eqref{eq:hybrid_friction_end} \quad & \text{Contact constraints} 
\end{align*}
\endgroup
Where the decision variables are indexed by finite element, $n$, collocation point, $p$, and contact phase, $s$.
Also included are problem specific constraints, such as start and end conditions, time constraints, control limits, and path constraints. 

The full HTO problem setup can now be defined, as summarized in Algorithm \ref{algo:sco_hto}. First, the contact sequence is extracted from the centroidal CIO result.
Second, the NLP variables and constraints are set up using the dynamics.
Finally, the body variables are copied over and interpolated using the centroidal CIO results. The leg joint angles are initialized using inverse kinematics at each step. By interpolating the centroidal CIO result, all variables in the HTO NLP can be initialized with a good initial guess. 

\begin{algorithm}[t]
\caption{SCO Stage 2 - Full-order HTO}
\label{algo:sco_hto}
\begin{algorithmic}

    \State Extract contact sequence from centroidal CIO solution
    \State Define HTO variables and constraints 
    \State Interpolate centroidal CIO solution and run inverse kinematics to initialize NLP variables

    \State Solve NLP2 w/ FE = 25 per mode, $3^{rd}$ order
    
\end{algorithmic}
\end{algorithm}

%%%%%%%%%%%%%%%%%%%%%%%%%%%%%%%%%%%%%%%%%%%%%%%%%%%%%%%%%%%%%%%%%%%%%%
\section{Performance Comparison} \label{sec:performance_comparison}

To determine when to use SCO instead of full-order CIO or HTO, three examples were tested. The flat ground example provides a simple proof of concept, demonstrating the advantages of using the stages within SCO, and shows that SCO may not be necessary when the optimal contact sequences is easily determined. The pit jump behavior provides an example where a standard HTO gait is not the most optimal contact sequence. The spatial quadruped example shows that SCO works for more complicated models.

%%%%%%%%%%%%%%%%%%%%%%%%%%%%%%%%%%%%%%%%%%%%%%%%%%%%%%%%%%%%%%%%%%%%%%
\subsection{Comparison Methods}

Here we detail the alternate trajectory optimization methods that are used to provide a comparison for SCO. Following the approach used within SCO to boost performance, an iterative optimization is used to refine the problem with tightening relaxation and mesh refinement, as well as taking advantage of first-order and third-order collocation.

%%%%%%%%%%%%%%%%%%%%%%%%%%%%%%%%%%%%%%%%%%%%%%%%%%%%%%%%%%%%%%%%%%%%%%
\subsubsection{Standard HTO}

For a standard HTO problem, the user provides a fixed contact sequence. In these comparisons, many contact sequences are tested to provide a fair comparison. Also, from our experience, HTO problems benefit greatly from a seeding approach, decreasing the solve times. Therefore, the problems are first solved with first-order collocation, and the result is then fed into the final third-order problem. The implementation details for the standard full-order HTO method can be seen in Algorithm~\ref{algo:standard_hto}.

\begin{algorithm}[t]
\caption{Standard Full-Order HTO}
\label{algo:standard_hto}
\begin{algorithmic}

    \State Define HTO variables and constraints
    %\If{Coordinate System = 2D}
    %    \State Iteration$\_$Cutoff = 500
    %\ElsIf{Coordinate System = 3D}
    %    \State Iteration$\_$Cutoff = 100
    %\EndIf
    
    \State Solve NLP2 w/ FE = 25 per mode, $1^{st}$ order
    \State Solve NLP2 w/ FE = 25 per mode, $3^{rd}$ order 
    
    \If{Use Terrain Map = True}
        \State Activate terrain map constraints
        \State Solve NLP2 w/ FE = 25 per mode, $3^{rd}$ order 
    \EndIf
    
\end{algorithmic}
\end{algorithm}

For the first-order collocation step, we choose to limit the number of iterations by the solver to 500 for the 2D problem and 100 for the 3D problem (tuned by hand to ensure that the seed is a reasonable first guess for the third-order step, but does not take up too much time). From our experience, running this step until convergence is not necessary, and only adds to the overall solve time of the algorithm. 

%%%%%%%%%%%%%%%%%%%%%%%%%%%%%%%%%%%%%%%%%%%%%%%%%%%%%%%%%%%%%%%%%%%%%%
\subsubsection{Full-Order CIO}

For the full-order CIO problem, the full robot dynamics are used, \eqref{eq:full_dynamics}, with the complementarity constraints,
\eqref{eq:non_penetration}--\eqref{eq:cio_relaxed}. This creates significant solving difficulty, and is notoriously finicky to use. Following \cite{posa2014direct,patel2019contact}, an iterative method of solving is used, summarized in Algorithm~\ref{algo:full_cio}. 
The optimization problem for full-order CIO is defined as:
\begingroup
\allowdisplaybreaks
\begin{align*} \label{problem:full_order_cio} 
\textbf{NLP3: Full-order}&\textbf{ CIO} \\
    \textbf{Minimize:} &
    \quad  \text{Full-order cost function \eqref{eq:hybrid_cost} }\\
    \textbf{With respect to:} \\
    q_{n,p}, \dot{q}_{n,p}, \ddot{q}_{n,p}  \quad & \text{Robot state, velocity, acceleration} \\
    a_{n,p}, \dot{a}_{n,p} \quad & \text{Foot position, velocity} \\
    \tau_{n} \quad & \text{Joint torques} \\
    \lambda_{n,p} \quad & \text{Ground reaction forces/impulses} \\
    \eta_{n,p} \quad & \text{Collocation slack variables} \\
    \alpha_{n,p}, \beta_{n,p}, \alpha^{'}_{n}, \beta^{'}_{n} \quad & \text{Complementarity slack variables} \\
    \textbf{Subject to:} \\
    \eqref{eq:euler}\text{ or }\eqref{eq:radau_end} \quad & \text{Collocation} \\
    \eqref{eq:full_dynamics} \quad & \text{Full order dynamics} \\
    \eqref{eq:fk_a} \quad & \text{Forward kinematics}\\
    \eqref{eq:non_penetration}-\eqref{eq:cio_relaxed} \quad & \text{Contact \& complementarity constraints} 
\end{align*}
\endgroup
Where the decision variables are indexed by finite element, $n$, and collocation point, $p$.
Also included are problem specific constraints, such as start and end conditions, time constraints, control limits, and path constraints. 

\begin{algorithm}[t] 
\caption{Standard Full-Order CIO}
\begin{algorithmic}

    \State Define full order CIO variables and constraints 

    \State Solve NLP3 w/ loose $\epsilon$, coarse FE, $1^{st}$ order
    \State Solve NLP3 w/ loose $\epsilon$, fine FE, $1^{st}$ order
    
    \If{Use Terrain Map = True}
        \State Activate terrain map constraints
        \State Solve NLP3 w/ loose $\epsilon$, fine FE, $1^{st}$ order
    \EndIf
    
    \If{Desired Collocation = $1^{st}$}
    
        \State Solve NLP3 w/ moderate $\epsilon$, fine FE, $1^{st}$ order
        \State Solve NLP3 w/ tight $\epsilon$, fine FE, $1^{st}$ order
        
    \ElsIf{Desired Collocation = $3^{rd}$}
    
        \State Solve NLP3 w/ loose $\epsilon$, fine FE, $3^{rd}$ order
        \State Solve NLP3 w/ moderate $\epsilon$, fine FE, $3^{rd}$ order
        \State Solve NLP3 w/ tight $\epsilon$, fine FE, $3^{rd}$ order
        
    \EndIf
    
\end{algorithmic}\label{algo:full_cio}
\end{algorithm}

%%%%%%%%%%%%%%%%%%%%%%%%%%%%%%%%%%%%%%%%%%%%%%%%%%%%%%%%%%%%%%%%%%%%%%
\subsubsection{HCIO}

The final method used in this comparison is the HCIO method \cite{mastalli2016hierarchical}, which uses a centroidal CIO stage to determine an initial seed for a full-order CIO stage. For a fair comparison, the iterative relaxation approach used in the other solving methods is implemented, see Algorithm~\ref{algo:hierarchical_cio}.

\begin{algorithm}[t] 
\caption{HCIO}
\begin{algorithmic}

    \State Call Centroidal CIO solve (Algorithm~\ref{algo:centroidal_cio_flat})
    \State Define full-order CIO variables and constraints 
    \State Interpolate centroidal CIO solution and run inverse kinematics to initialize NLP variables
    
    \If{Use Terrain Map = True}
        \State Activate terrain map constraints
    \EndIf
    
    \State Solve NLP3 w/ loose $\epsilon$, fine FE, $1^{st}$ order
    
    \If{Desired Collocation = $1^{st}$}
    
        \State Solve NLP3 w/ moderate $\epsilon$, fine FE, $1^{st}$ order
        \State Solve NLP3 w/ tight $\epsilon$, fine FE, $1^{st}$ order
        
    \ElsIf{Desired Collocation = $3^{rd}$}
    
        \State Solve NLP3 w/ loose $\epsilon$, fine FE, $3^{rd}$ order
        \State Solve NLP3 w/ moderate $\epsilon$, fine FE, $3^{rd}$ order
        \State Solve NLP3 w/ tight $\epsilon$, fine FE, $3^{rd}$ order
        
    \EndIf
    
\end{algorithmic}\label{algo:hierarchical_cio}
\end{algorithm}

%%%%%%%%%%%%%%%%%%%%%%%%%%%%%%%%%%%%%%%%%%%%%%%%%%%%%%%%%%%%%%%%%%%%%%
\subsection{Implementation Specifics}

All optimization problems have been set up in Python, using the Pyomo optimization package~\cite{hart2011pyomo,hart2017pyomo}.  Optimizations are solved using IPOPT~\cite{wachter2006implementation}, with HSL linear solvers~\cite{hsl}, and run on an Intel(R) Core(TM) i9-9900k CPU @ 3.60GHz.

Problems with terrain maps are implemented using a sigmoid approximation to a step function as follows:
\begin{equation} \label{eq:sigmoid}
    h_{0,i} = \sum_k \frac{\zeta_k}{1 + \exp(-\psi_ka_{x,i} - \xi_k)}
\end{equation}
where $h_{0,i}$ is the ground height, $\zeta_k$ is the step magnitude, $\psi_k$ is the slope, $a_{x,i}$ is the $x$-coordinate of the $i^{th}$ contact point, and $\xi_k$ is the $x$-offset of the step. For each terrain map, a list of $\zeta$, $\psi$, and $\xi$ are made for each of the $k$ sigmoids. To use this terrain map with the constraints introduced in the NLP setup, the sum of sigmoids is differentiated to determine the angle of the ground w.r.t. to horizontal. This angle is then applied to all constraints and variables in the NLP.

The relaxation value, $\epsilon$, was defined on a scale ranging from loose ($10^1$) to moderate ($10^{-1}$) to tight ($10^{-3}$). Similarly, for the number of FE, the best results came with coarse ranging from 12-25, and fine ranging from 100-200, depending upon the behavior.

%%%%%%%%%%%%%%%%%%%%%%%%%%%%%%%%%%%%%%%%%%%%%%%%%%%%%%%%%%%%%%%%%%%%%%
\subsection{Flat Ground Walking Task}

In the flat ground walking task for the planar biped robot, the robot is required to move 2.5m forward while minimizing the integral of the total squared joint torques. Here the optimal contact sequence is a walking gait comprised of instantaneous switches between left and right legs. However, the exact number of steps must be found by either using a step length heuristic or running multiple HTO trials. The cost and timing results for each approach are listed in Table~\ref{Flat_Ground_Results}. % and in Fig.~\ref{FlatComp}. 

% Old results (Grad school machine)
% \begin{table}[t]
% \caption{Results for planar biped robot examples, flat ground and pit jump, minimizing sum of motor torques squared.}
% \begin{center}
% \label{Flat_Ground_Results}
% \begin{tabular}{c c c c}
% \hline
% Terrain & Method & Cost ((Nm)$^2$s) & Time (min:sec)\\
% \hline
% Flat & Full-Order CIO & 1806 & 45:08 \\
% Flat & HCIO & 1583 & 28:30 \\
% Flat & HTO (best case) & 1508 & 2:48 \\
% Flat & HTO (SCO Gait) & 1758 & 3:22 \\
% Flat & SCO & 1758 & 1:46 \\
% \hline
% Pit & Full-Order CIO & 2012 & 41:40 \\
% Pit & HCIO & 2346 & 60:29 \\
% Pit & HTO (best case) & 2120 & 3:44 \\
% Pit & HTO (SCO Gait) & XXXX & XXXX \\
% Pit & SCO & 2048 & 2:10 \\
% \hline
% \end{tabular}
% \end{center}
% \end{table}

\begin{comment}
\begin{figure}[t]
\begin{center}
\includegraphics[width=0.95\columnwidth]{FlatComp.png}
\end{center}
\caption{Results for flat ground example. Top: Full-order CIO result. Second: Most optimal HTO gait. Third: SCO centroidal CIO solution. Bottom: SCO result. For this flat ground example, iteration over multiple HTO gaits finds the most optimal solution. Both full-order CIO and SCO fall behind in optimal by around 20$\%$. Full-order CIO solve times are an order of magnitude longer, however, SCO is able to autonomously determine the contact sequence and the trajectory in a slightly faster time than HTO.}
\label{FlatComp} 
\end{figure}
\end{comment}

Given the optimal 10 mode walking gait, HTO finds the lowest cost result, with the other methods all falling within 20\% of this best HTO cost. However, this contact sequence was determined after multiple iterations of HTO with different contact sequences to narrow in on the most optimal result. In practice, the user would need to run trials on various gait cycles, which can be time consuming. 

% New results (new imaged machine)
\begin{table}[t]
\vspace{1em}
\caption{Results for planar biped robot examples, flat ground and pit jump, minimizing sum of motor torques squared.}
\begin{center}
\label{Flat_Ground_Results}
\begin{tabular}{c c c c}
\hline
Terrain & Method & Cost ((Nm)$^2$s) & Time (min:sec)\\
\hline
Flat & Full-Order CIO & 1812 & 25:07 \\
Flat & HCIO & 1606 & 41:16 \\
Flat & HTO (walking gait) & \textbf{1508} & 2:14 \\
Flat & HTO (SCO Gait) & 1758 & \textbf{1:02} \\
Flat & SCO & 1758 & 1:44 \\
\hline
Pit & Full-Order CIO & 2148 & 61:07 \\
Pit & HCIO & \textbf{1998} & 66:41 \\
Pit & HTO (running gait) & 2120 & 7:58 \\
Pit & HTO (SCO Gait) & 2038 & 4:21 \\
Pit & SCO & 2038 & \textbf{1:54} \\
\hline
\end{tabular}
\end{center}
\end{table}

Looking further into the SCO timing, the CIO step took 60s and the HTO step took 43s. When running HTO from scratch with the SCO selected gait, we arrive at the same final gait and cost, but the execution time is higher than the HTO stage of SCO. This highlights the benefit of the staged approach in both finding a good (though maybe not globally optimal) contact sequence as well as a good initial trajectory seed with that contact sequence.

HCIO finds a solution very close in optimality to the most optimal HTO gait. However, the final result is somewhat sporadic, making and breaking contact intermittently during the trajectory when not necessary. This highlights that using full-order CIO for this final pass of the algorithm can still result in a shaky final solution.

Overall, despite not being as optimal as HTO, SCO was able to autonomously determine a contact sequence for the situation and do so in a similar time as a standard HTO problem, gaining CIO abilities while keeping HTO speed.

%%%%%%%%%%%%%%%%%%%%%%%%%%%%%%%%%%%%%%%%%%%%%%%%%%%%%%%%%%%%%%%%%%%%%%
\subsection{Pit Jump Task}

\begin{figure}[t]
\begin{center}
\vspace{1em}
\includegraphics[width=0.7\columnwidth]{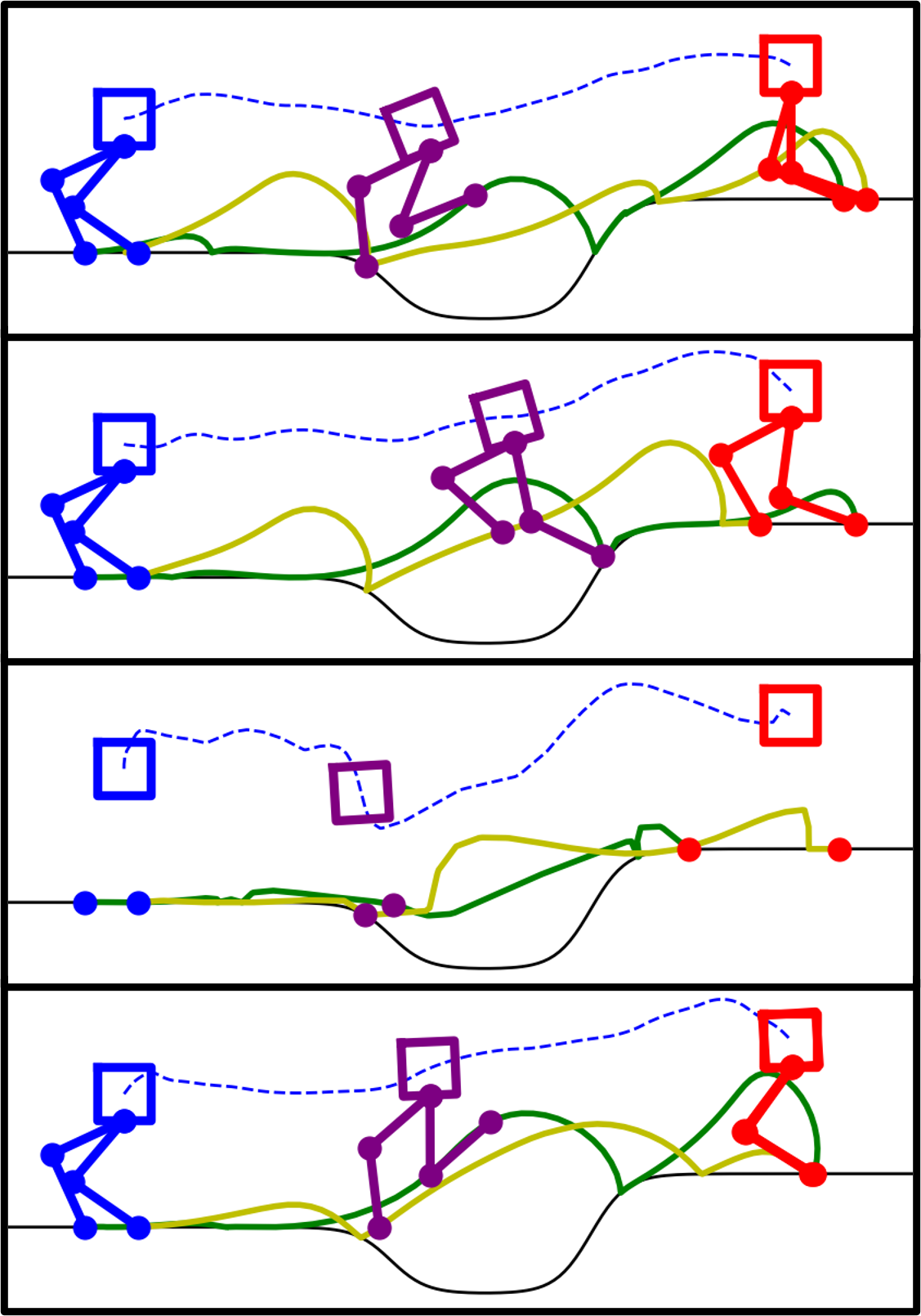}
\end{center}
\vspace{-1em}
\caption{Results for pit jump. Top: Full-order CIO. Second: Most optimal HTO running gait. Third: SCO centroidal CIO. Bottom: SCO result. For this example, HCIO finds the most optimal solution, followed closely by SCO. HCIO solve times are an order of magnitude longer, however, SCO is able to autonomously determine the contact sequence and the trajectory around four times as fast as standard HTO.}
\label{PitJumpComp} 
\end{figure}

In this task, the planar biped robot is required to run and jump over a pit, Fig.~\ref{PitJumpComp}. A varying terrain map increases the complexity of the optimal contact sequence, unlike most flat ground results that have an optimal contact sequence of a standard walking gait. For HTO, the most versatile gait to use on terrain maps for the planar biped robot is the running gait (alternating legs with a flight phase in between). In this situation, full-order CIO and SCO should be able to find more optimal results, as the flight phase is not necessary over the flat portions of the terrain. The cost and timing results for each approach are listed in Table~\ref{Flat_Ground_Results}.

While HCIO finds the most optimal result for this task, SCO finds a solution only 2$\%$ less optimal in significantly less time. SCO solved faster and more optimally than all variations of the standard running HTO gaits. In this case SCO was much faster than HTO even with the same contact sequence, as the simplified earlier stages helped guide the search. This shows SCO's ability to quickly find solutions that are more optimal than standard HTO on more complicated tasks. 

%%%%%%%%%%%%%%%%%%%%%%%%%%%%%%%%%%%%%%%%%%%%%%%%%%%%%%%%%%%%%%%%%%%%%%
\subsection{Spatial Quadruped Walking Spin}

\begin{figure*}[t]
\begin{center}
\includegraphics[width=1.0\textwidth]{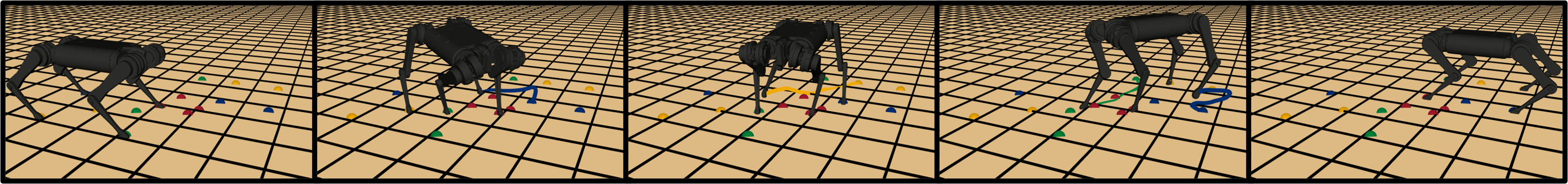}
\end{center}
\vspace{-1.5em}
\caption{SCO result for spatial quadruped robot performing a walking spin, minimizing sum of motor torques squared. Footholds and current swing leg trajectories are shown to help visualize the complex behavior.}
\label{Spin_Step_Back_Figure} 
\end{figure*}

In order to show the potential of SCO on more complicated systems, the algorithm was applied to a spatial, 18 DOF quadruped. Due to significantly more possibilities, the optimal contact sequence is difficult to find. This results in users having to perform gait studies, and apply these cyclical gaits to tasks that would be better suited for a unique contact sequence. In this trial, the quadruped is required to perform a walking spin, moving 0.5m forward and 0.5m to the left, while rotating 180 degrees, Fig.~\ref{Spin_Step_Back_Figure}. The results, Table~\ref{Spin_Step_Back_Results}, show that again SCO and HTO are much faster than CIO, and HTO is heavily reliant on the quality of the user input gait.

% New machine results
\begin{table}[t]
\caption{Results for spatial quadruped robot performing a walking spin, minimizing sum of motor torques squared.}
\begin{center}
\label{Spin_Step_Back_Results}
\begin{tabular}{c c c}
\hline
Method & Cost ((Nm)$^2$s) & Time (hr:min)\\
\hline
Full-Order CIO & FAIL & 36:00 \\
HCIO & 166 & 20:18 \\
HTO (standing trot) & \textbf{129} & \textbf{0:37} \\
HTO (flying trot) & 602 & 1:43 \\
SCO & 270 & 2:08 \\
\hline
\end{tabular}
\end{center}
\end{table}

% Old machine results
% \begin{table}[t]
% \caption{Results for spatial quadruped robot performing a walking spin, minimizing sum of motor torques squared.}
% \begin{center}
% \label{Spin_Step_Back_Results}
% \begin{tabular}{c c c}
% \hline
% Method & Cost ((Nm)$^2$s) & Time (min:sec)\\
% \hline
% Full-Order CIO & 129.4 & 790:15 \\ need to run these again
% HCIO & xxxx & xxxx \\ need to run these again
% HTO (best case) & 105.1 & 104:56 \\t
% SCO & 168.2 & 95:46 \\
% \hline
% \end{tabular}
% \end{center}
% \end{table}

Comparing the timing results across different methods, full-order CIO and HCIO took dramatically longer than HTO and SCO. HTO with a standing trot gait is the most optimal in terms of cost and time overall. However, SCO's result is almost twice as optimal as HTO, with one of the most commonly used gaits, the flying trot, while taking a similar amount of time to solve. We also note that full-order CIO failed to converge for this trial after over 36 hours, and both full-order CIO and HCIO had some inconsistencies converging for other complicated spatial quadruped tasks.

%%%%%%%%%%%%%%%%%%%%%%%%%%%%%%%%%%%%%%%%%%%%%%%%%%%%%%%%%%%%%%%%%%%%%%
\section{Conclusion} \label{sec:conclusion}

This paper presents a new approach to trajectory optimization with unknown contact sequences -- Staged Contact Optimization (SCO). Results show SCO's ability to quickly find feasible contact schedules and solve for refined, full-order trajectory solutions. SCO retains the flexibility of contact scheduling from CIO, but without the long solve times. By fixing the contact modes of the CIO result and using a good seed, SCO is able to use HTO to efficiently solve for the full-order trajectory solution. SCO had similar or faster solution times than HTO without requiring a fixed mode schedule. In the examples where the optimal mode sequence is difficult to guess, SCO was able to find a more optimal solution than HTO.

This approach can be improved by making changes to the algorithm for either of the constituent stages. Faster NLP solvers, e.g.~\cite{howell2019altro}, that specialize in CIO or HTO problems will also benefit SCO. There are still many parameters to choose in the SCO solution strategy, such as the centroidal cost function, the number of solves, the relaxation $\epsilon$, the number of finite elements, etc. Further work will analyze these parameters to better understand how they affect solution time and optimality, as well variations for different problems.

%%%%%%%%%%%%%%%%%%%%%%%%%%%%%%%%%%%%%%%%%%%%%%%%%%%%%%%%%%%%%%%%%%%%%%
\section*{Acknowledgments}
The authors would like to thank Stacey Shield and Amir Patel for their help with implementation details~\cite{shieldpyomo}.

%%%%%%%%%%%%%%%%%%%%%%%%%%%%%%%%%%%%%%%%%%%%%%%%%%%%%%%%%%%%%%%%%%%%%%


\bibliographystyle{IEEEtran}

\bibliography{references}

\end{document}